\pdfoutput=1

\documentclass[11pt]{article}

\usepackage[preprint]{acl}

\usepackage{times}
\usepackage{latexsym}

\usepackage[T1]{fontenc}

\usepackage[utf8]{inputenc}

\usepackage{microtype}

\usepackage{inconsolata}

\usepackage{graphicx}

\usepackage{amsmath}

\title{Uncovering Cross-Linguistic Disparities in LLMs using Sparse Autoencoders}

\author{
 \textbf{Richmond Sin Jing Xuan\textsuperscript{1}},
 \textbf{Jalil Huseynov \textsuperscript{1}},
 \textbf{Yang Zhang\textsuperscript{1}}
\\
\\
 \textsuperscript{1} National University of Singapore
}

\begin{document}
\maketitle
\begin{abstract}
Multilingual Large Language Models (LLMs) exhibit strong cross-linguistic generalization, yet medium-to-low resource languages underperform in common benchmarks (ARC-C, MMLU, HellaSwag). We analyze activation patterns in Gemma-\(2\)-\(2\)B across all \(26\) residual layers and \(10\) languages: Chinese (zh), Russian (ru), Spanish (es), Italian (it); medium-to-low resource languages: Indonesian (id), Catalan (ca), Marathi (mr), Malayalam (ml), Hindi (hi), with English (en) as the reference—using Sparse Autoencoders (SAEs), revealing systematic disparities—medium-to-low resource languages receive up to \(26.27\%\) lower activations in early layers, persisting at \(19.89\%\) in deeper layers. To address this, we apply activation-aware fine-tuning via LoRA, leading to substantial activation gains (e.g., Malayalam \(87.69\%\), Hindi \(86.32\%\)) while maintaining English retention (\(\sim91\%\)). Post-fine-tuning, benchmarks show modest but consistent improvements, highlighting activation alignment as a crucial factor in multilingual LLM performance.
\end{abstract}

\section{Introduction}
Large Language Models (LLMs), such as GPT-\(4\) \citep{openai2023}, Gemma \citep{gemma2024}, and LLaMA \(3\) \cite{meta2024}, demonstrate strong multilingual capabilities in zero-shot and few-shot settings. These models leverage shared cross-linguistic representations, enabling them to process both high- and low-resource languages. However, multilingual training data remains highly imbalanced, with English dominating most pre-training corpora. OpenAI (\(2023\)) reported that \(92.65\%\) of GPT-\(3\)’s training tokens were English, while Meta (\(2024\)) disclosed that non-English data makes up just over \(5\%\) of LLaMA \(3\)’s dataset. This raises concerns about whether LLMs are disproportionately optimized for high-resource languages \citep{blasi2021}.\\
\noindent To investigate this multilingual performance gap, we leverage Sparse Autoencoders (SAEs) \citep{cunningham2023sparse}—an interpretability tool designed to analyze internal neuron activations in LLMs. SAEs have been widely used in mechanistic interpretability research \citep{bricken2023monosemanticity, templeton2024scaling, kissane2024interpreting} to extract structured, human-interpretable features from complex activations. Unlike embeddings, which measure representational similarity, SAEs provide direct insights into how LLMs process different languages at the neuron activation level. A key recent advancement in this space is Gemma Scope \citep{lieberum2024}, an open-source framework that applies SAEs to Gemma-\(2\)-\(2\)B \citep{gemma2_2024}, allowing for precise layer-wise analysis of neuron activations linked to linguistic properties.\\
\noindent
Building on this framework, we investigate whether LLMs exhibit systematic activation disparities between high-resource and medium-to-low resource languages, despite similar embeddings. Using SAEs on Gemma-\(2\)-\(2\)B, we conduct a layer-wise activation analysis across multiple languages to assess whether activation differences contribute to the multilingual performance gap.\\
\noindent
Our key contributions are as follows: 

1. We analyze embedding similarity across languages and demonstrate that despite appearing to capture all languages similarly, LLMs exhibit significant performance disparities in real-world tasks, as evidenced by lower scores for medium-to-low resource languages on Okapi's evaluation framework \citep{lai2023okapi}. 

2. We conduct a layer-wise SAE activation study, showing that medium-to-low resource languages exhibit significantly lower activation values compared to high-resource languages, particularly in the early layers of processing. 

3. We demonstrate that fine-tuning with activation alignment reduces activation disparities and yields modest improvements in model performance, providing a foundation for future work in improving multilingual representations in LLMs.\\
\noindent
Our findings challenge the assumption that embedding similarity guarantees equal linguistic understanding and highlight the importance of activation-based interpretability techniques in multilingual NLP. By identifying and quantifying cross-linguistic activation disparities, we contribute to the broader discourse on multilingual fairness and propose new strategies for improving LLM performance on medium-to-low resource languages.
\section{Methodology}
In this section, we outline our methodology for investigating multilingual representation in LLMs. We first introduce the dataset and preprocessing pipeline. Next, we assess whether LLMs encode different languages equally. To evaluate whether these similarities translate to actual performance, we conduct multilingual benchmark tests using Okapi’s framework. Finally, we perform SAE-based activation analysis to uncover deeper activation disparities across languages and apply fine-tuning with activation alignment to mitigate activation disparities and improve model performance.

\subsection{Data Collection and Preprocessing}
\label{2.1}
We use the Neuronpedia \citep{bloom2024saetrainingcodebase} dataset, which contains \(16,384\) interpretable neuron features across all \(26\) layers of Gemma-\(2\)-\(2\)B. Each feature represents a learned linguistic property identified via SAE decomposition. To construct our dataset, we first apply tokenization constraints by segmenting each English phrase using a ±\(3\) token window around the most activated token, preserving contextual relevance while ensuring efficient processing. \\
\noindent
We then extract top-activated English phrases exceeding an \(80\%\) activation threshold relative to the highest activation in that feature and translate them into nine languages (\textit{zh}, \textit{ru}, \textit{es}, \textit{it}, \textit{id}, \textit{ca}, \textit{mr}, \textit{ml}, \textit{hi}) using Helsinki-NLP models \citep{TiedemannThottingal:EAMT2020, tiedemann2023democratizing}. This yields a multilingual dataset of approximately six million phrases.

\subsection{Embedding-Based Activation Analysis}  
We analyze cross-linguistic representations in Gemma-\(2\)-\(2\)B by computing cosine similarity between residual activations of English phrases and their translations across nine languages. 
In our collected dataset, each phrase is associated with different feature indices. We use data from feature indices sampled at intervals of \(16\), selecting $\{16i \mid i \in \{0, \dots, 999\}\}$ from \(16,384\) available indices to balance efficiency and representativeness.\\
\noindent
To obtain fixed-size activation vectors, we apply mean pooling over token activations per phrase. For each layer, we compute the cosine similarity between activation vectors of translated phrases and their English counterparts and then average these values across all phrases. This yields per-layer similarity scores that quantify embedding alignment across languages, further discussed in Section~\ref{3.1}.

\subsection{Multilingual Benchmark Evaluation}
Using Okapi’s framework, we evaluate LLMs accuracy on multilingual MMLU \citep{hendrycks2020mmlu}, HellaSwag \citep{zellers2019hellaswag}, and ARC-Challenge \citep{clark2018arc}, comparing high-resource and medium-to-low resource languages to assess alignment between embeddings and actual performance.

\subsection{SAE-Based Activation Analysis}
To go beyond embeddings, we analyze activation-level disparities using SAEs trained on Gemma-\(2\)-\(2\)B. Given a phrase \( P \) in language \( L \), we extract its activation vector at layer \( l \):
\[
A^L_l = \text{SAE}(P, l)
\]
For each feature index, multiple phrases contribute activation values. We compute the mean activation per index by averaging across all associated phrases, allowing a direct comparison of activation magnitudes at each index.

\subsection{Cross-Linguistic Activation Disparities}
Finally, we quantify activation differences between high-resource and medium-to-low resource languages. For each layer \( l \), we compute the percentage activation gap between the two groups:
\[
\Delta A_l = \frac{\bar{A}_{\text{high}} - \bar{A}_{\text{med-low}}}{\bar{A}_{\text{high}}} \times 100
\]
where \( \bar{A}_{\text{high}} \) and \( \bar{A}_{\text{med-low}} \) represent the mean activation scores for high-resource and medium-to-low resource languages, respectively. By analyzing \( \Delta A_l \) across all layers, we uncover whether LLMs activate differently for different language groups, highlighting potential biases in how features are learned and processed.

\subsection{Fine-Tuning for Improved Model Performance}
To enhance multilingual performance, we apply LoRA-based fine-tuning \citep{hu2021lora} with an activation alignment objective. The loss function is defined as:
\[
f(u_l, v_l) = |u_l - v_l| + \alpha \cdot ||u_l - u_{\text{orig},l}||^2
\]
where \( u_l \) represents the activation of English at layer \( l \), \( v_l \) represents the activation of a corresponding language, and \( u_{\text{orig},l} \) is the original English activation recorded in Neuronpedia. The first term minimizes activation gaps between English and the target language, while the second term ensures English activations remain stable during fine-tuning. Fine-tuning results are analyzed in Section~\ref{3.3}.
\section{Experiments}
We conduct experiments using the Gemma-\(2\)-\(2\)B model and analyze cross-linguistic activation patterns with Gemma Scope, an interpretability framework based on Sparse Autoencoders (SAEs).\\
\noindent
Section~\ref{3.1} examines the gap between embedding similarity and benchmark performance, highlighting multilingual disparities. Section~\ref{3.2} analyzes activation differences across languages using SAEs. Section~\ref{3.3} evaluates fine-tuning with activation alignment to reduce activation gaps and enhance cross-linguistic performance.

\subsection{Embedding-Based Performance Disparities}
\label{3.1}

\begin{figure}[t]
    \centering
    \includegraphics[width=1\linewidth]{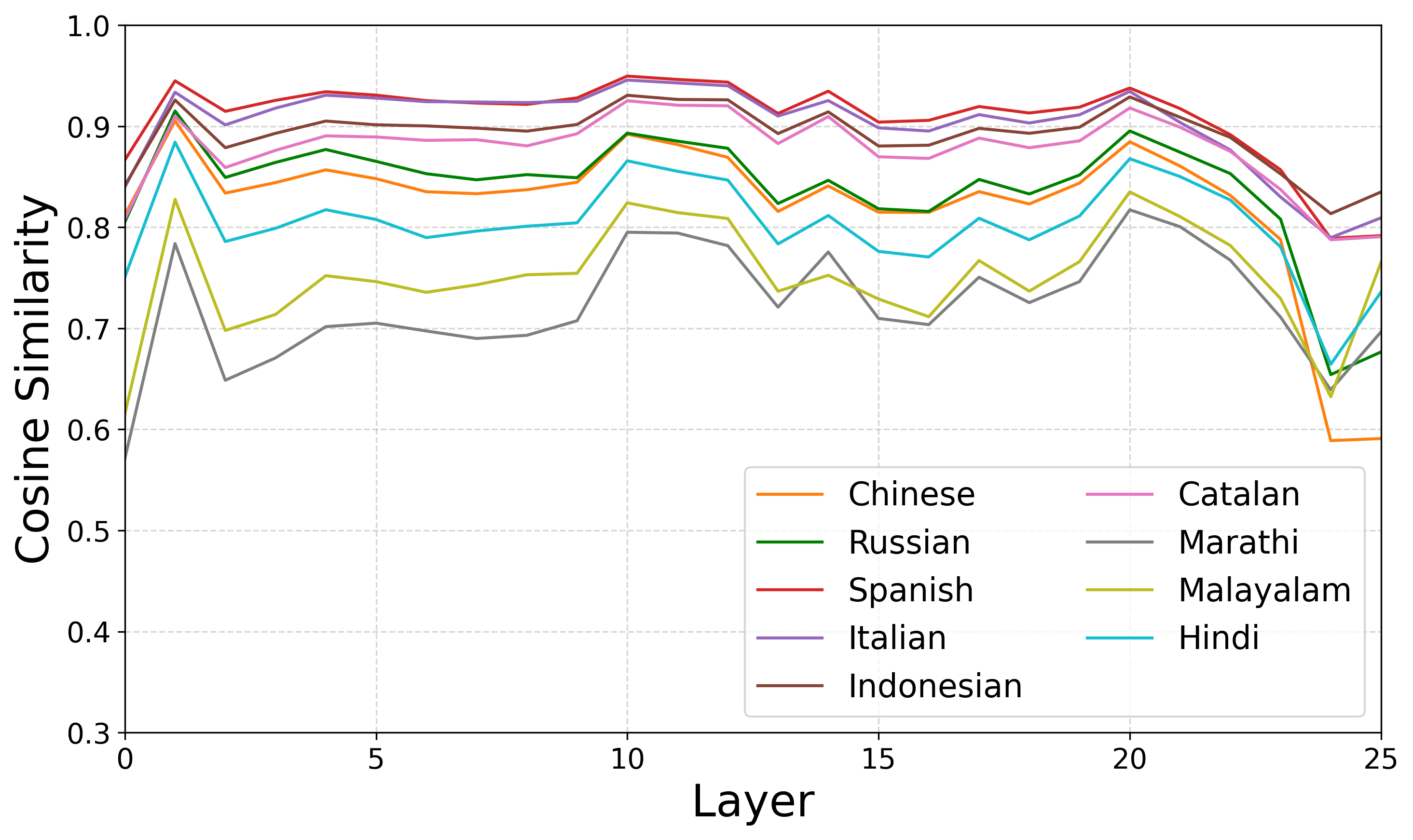}
    \caption{Cross-linguistic embedding similarity between English and nine languages across all layers.}
    \label{fig:cosine_similarity}
\end{figure}

\begin{figure}[t]
    \centering
    \includegraphics[width=1\linewidth]{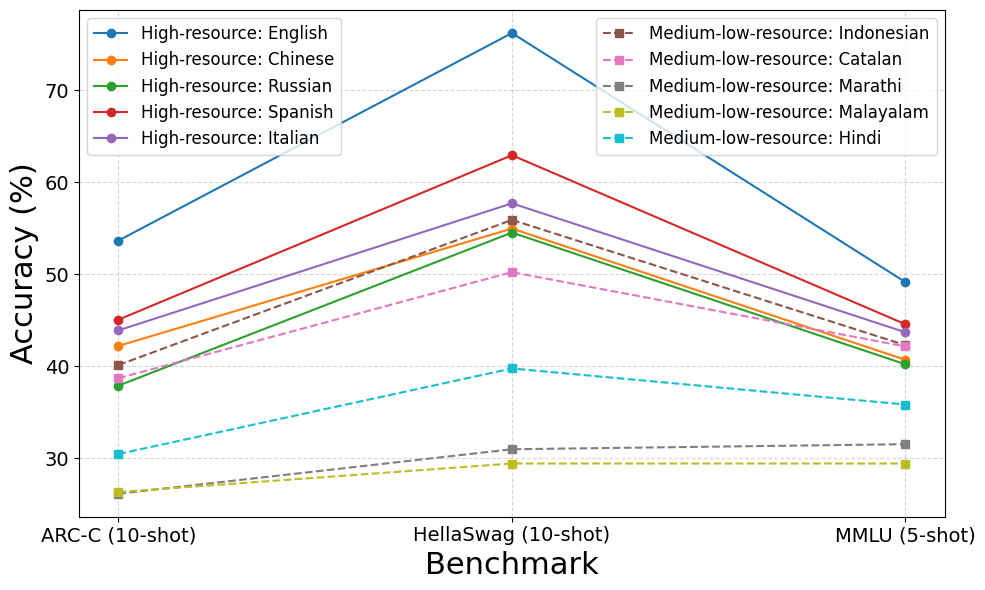}
    \caption{Performance between high-resource and medium-to-low resource languages across all benchmarks.}
    \label{fig:benchmark_performance}
\end{figure}

\noindent
Embedding similarity is often used as a proxy for multilingual generalization \citep{dubossarsky2020spectra}, but our analysis shows it does not always correlate with strong benchmark performance.\\
\noindent
We compute cosine similarity between English residual activations and their translations in nine languages using Gemma-\(2\)-\(2\)B. As shown in Figure~\ref{fig:cosine_similarity}, similarity declines in the initial and final layers, aligning with their encoding and text generation roles \citep{zhang2024layerimportance}. In intermediate layers, most languages—except Malayalam and Marathi—maintain similarity above \(0.8\), while all remain above \(0.7\), aside from slight dips in some middle layers. However, despite strong embedding alignment, benchmark results (Figure~\ref{fig:benchmark_performance}) reveal substantial performance gaps.\\
\noindent
While residual activation similarities remain high, ARC-Challenge (\(10\)-shot) scores drop from \(53.67\%\) (English) to \(26.16\%\) (Marathi) and \(26.34\%\) (Malayalam), and HellaSwag (\(10\)-shot) shows a \(46.79\%\) gap between English and Malayalam, indicating surface-level alignment does not guarantee equitable activation dynamics. Furthermore, medium-to-low resource languages consistently underperform across all benchmarks, particularly ARC-Challenge and MMLU, where Hindi, Marathi, and Malayalam score significantly lower despite high embedding similarity \citep{robinson2023chatgpt}. 

\subsection{SAE-Based Activation Analysis}
\label{3.2}
\begin{figure}[ht]
\centering
\includegraphics[width=1\linewidth]{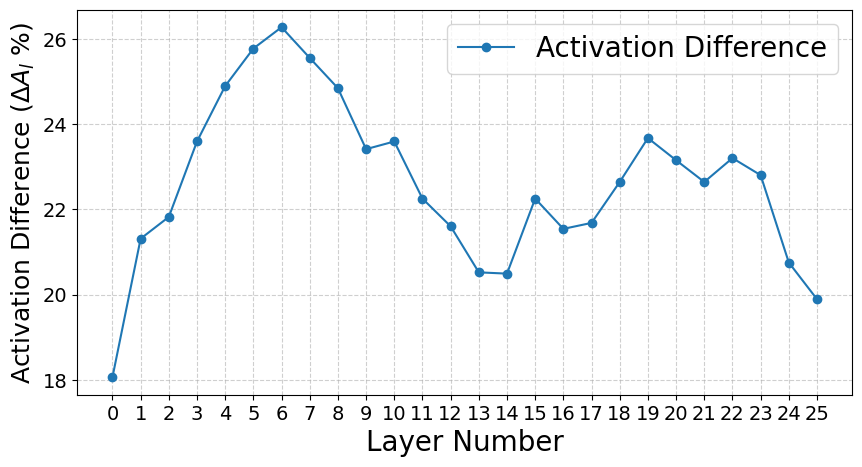}
\caption{Differences in activation between high-resource and medium-to-low resource languages across layers.}
\label{fig:activation_differences}
\end{figure}

To investigate performance disparities, we analyze cross-linguistic activation patterns using Sparse Autoencoders (SAEs). Figure~\ref{fig:activation_differences} visualizes activation differences across \(26\) layers.\\

\noindent
These differences lead to three key observations:
1. Peak disparities in early layers. Activation gaps are largest at layer \(6\) (\(26.27\%\)), suggesting weaker encoding of lower resource languages, consistent with prior findings on syntactic and semantic feature activation \citep{nanda2023}.\\
\noindent
2. Reduction but persistence of disparities in deeper layers. Activation gaps decrease to \(19.89\%\) at layer \(25\), suggesting that deeper layers rely more on shared representations rather than language-specific features \citep{dufter2020necessary, paul2024ircoder, pires2019}.\\
\noindent
3. Activation gaps correlate with performance. Medium-to-low resource languages show consistent \(20\%\) lower activations, mirroring weaker common benchmark results (see Figure~\ref{fig:benchmark_performance}), aligning with evidence that activation disparities hinder multilingual generalization \citep{conneau2020}. Figures~\ref{fig:figure_7}, \ref{fig:figure_8}, and \ref{fig:figure_9} show that activation gap is indeed negatively correlated with benchmark performance.

\subsection{Fine-Tuning for Improved Model Performance}
\label{3.3}
Fine-tuning improves activation alignment across languages, leading to measurable gains in both activation consistency and benchmark performance. 
\begin{figure}[t]
    \centering
    \includegraphics[width=1\linewidth]{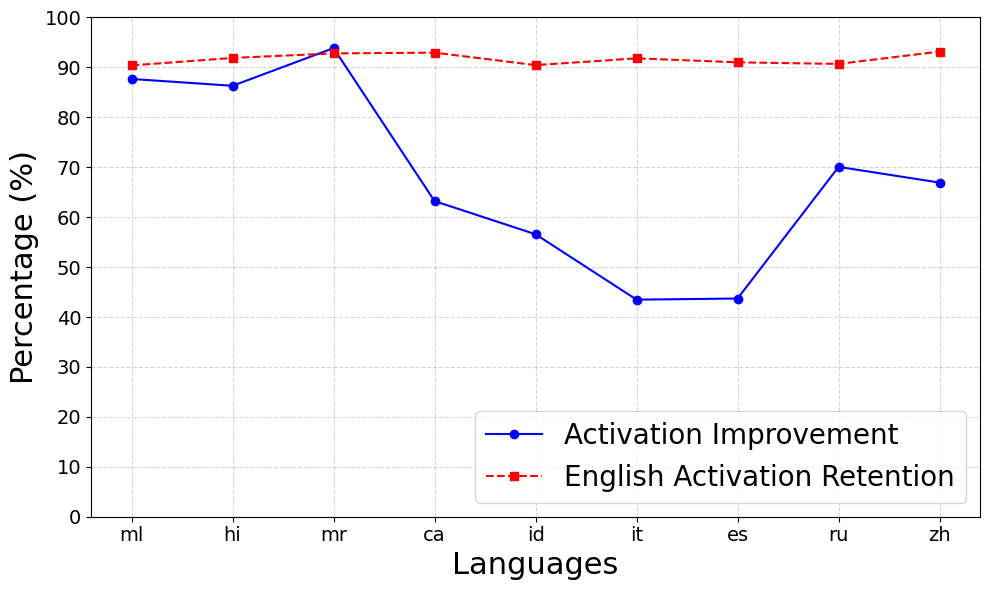}
    \caption{Activation improvements and English retention compared to the original score across all languages in layer \(20\).}
    \label{fig:all_lang_performance}
\end{figure}
Fine-tuning yields two key outcomes:\\
\noindent
1. Activation gains across languages with English stability. Fine-tuning increased activations across all ten languages in layer \(20\) (Figure~\ref{fig:all_lang_performance}), with the highest improvements in Marathi (\(93.87\%\)), Malayalam (\(87.69\%\)), and Hindi (\(86.32\%\)). Even high-resource languages like Russian (\(70.08\%\)) and Chinese (\(66.93\%\)) saw substantial gains. English activations remained stable (\(\geq90\%\) retention), ensuring fine-tuning did not distort learned representations. This trend generalizes across layers, as shown by the reduction of activation gaps in Malayalam (see Appendix Figure~\ref{fig:malayalam_all_layers}).\\
\noindent
2. Improved benchmark performance. Fine-tuning resulted in a \(1.44\)-point increase in ARC-C (Malayalam) accuracy, translating to a \(5.47\%\) improvement, demonstrating the positive impact of activation alignment on downstream tasks \citep{conneau2020, su2024finetuning}. Additional benchmark results (Malayalam) for other tasks can be found in Appendix Figure~\ref{fig:benchmark_results_malayalam}.

\section{Limitations}
Our study has three main limitations. First, while Helsinki-NLP models enable efficient large-scale translation, they may introduce errors that misalign activation patterns, particularly for medium-to-low resource languages, introducing bias in cross-lingual analysis \citep{agrawal2024translation}. 
Second, while fine-tuning reduced activation disparities, benchmark performance improvements remained modest. This suggests that activation alignment alone is insufficient, as multilingual LLMs do not fully converge to shared representations across languages \citep{zeng2024converging}.
Third, our findings are based on Gemma-\(2\)-\(2\)B, and it remains unclear whether similar activation disparities exist across other architectures. Multilingual activation studies indicate that disparities may vary across models, warranting further investigation \citep{liu2024unraveling}.

\section{Conclusion}
This study reveals that LLMs exhibit systematic activation disparities across languages, despite embedding similarity. Medium-to-low resource languages consistently receive lower activations, which correlate with weaker performance on multilingual benchmarks. While fine-tuning reduces activation gaps, its effect on task performance remains limited, highlighting the need for more effective adaptation strategies. Future work should explore refining fine-tuning techniques to better translate activation alignment into performance gains and extend this analysis to other LLMs and multilingual datasets to further improve language equity in LLMs.
\newpage

\bibliography{custom}
\newpage
\appendix
\section{Appendix}

\subsection{Related Works}
Sparse Autoencoders (SAEs) have been widely adopted in mechanistic interpretability for decomposing neural activations into monosemantic latent features. However, recent studies have raised concerns about their ability to recover truly independent and disentangled features. \citet{till_2024_saes_true_features} argues that $\ell_1$ regularization, which penalizes activations to enforce sparsity, may cause neurons to encode frequently co-occurring feature combinations rather than distinct, independent features. As a result, some rare latent features may not be learned at all. Furthermore, \citet{anders_etal_2024_composedtoymodels_2d} demonstrates that SAEs consistently learn composed features rather than the true underlying features, regardless of learning rate or $\ell_1$ coefficient used in training. These findings suggest that SAEs may distort learned representations, potentially limiting their interpretability. \\
\noindent 
While these concerns focus on the properties of SAEs themselves, our study suggests that the observed activation gaps may be a result of the model struggling to generalize in medium-to-low resource languages, rather than a fundamental issue with SAE representations. Specifically, we find that activation vectors appear well-aligned across languages at the embedding layer, but a significant gap emerges at the SAE layer between high-resource and medium-to-low resource languages. This aligns with performance disparities observed in benchmark tasks such as ARC-C, HellaSwag, and MMLU, where model accuracy declines for medium-to-low resource languages. Our results suggest that the model itself may struggle to develop robust representations for these languages, which is reflected in the activation patterns extracted by SAEs. This finding underlines the broader challenge of ensuring high-quality, language-agnostic representations in multilingual models.

\subsection{Preprocessing and Data Preparation}

\subsubsection{Gemma-2-2B and Gemma Scope Local Deployment}

We deploy Gemma-\(2\)-\(2\)B and Gemma Scope locally to enable efficient processing and analysis of neural activations across multiple languages. This setup provides local inference capabilities, allowing structured analysis of neuron activations without cloud-based constraints.

\paragraph{Layers Processed:} Gemma-\(2\)-\(2\)B consists of \(26\) layers, each containing \(16,384\) indices, representing distinct learned features. The layers processed in this study are:

\begin{itemize}
    \item \texttt{layer\_0/width\_16k/average\_l0\_105}
    \item \texttt{layer\_1/width\_16k/average\_l0\_102}
    \item \texttt{layer\_2/width\_16k/average\_l0\_141}
    \item \texttt{layer\_3/width\_16k/average\_l0\_59}
    \item \texttt{layer\_4/width\_16k/average\_l0\_124}
    \item \texttt{layer\_5/width\_16k/average\_l0\_68}
    \item \texttt{layer\_6/width\_16k/average\_l0\_70}
    \item \texttt{layer\_7/width\_16k/average\_l0\_69}
    \item \texttt{layer\_8/width\_16k/average\_l0\_71}
    \item \texttt{layer\_9/width\_16k/average\_l0\_73}
    \item \texttt{layer\_10/width\_16k/average\_l0\_77}
    \item \texttt{layer\_11/width\_16k/average\_l0\_80}
    \item \texttt{layer\_12/width\_16k/average\_l0\_82}
    \item \texttt{layer\_13/width\_16k/average\_l0\_84}
    \item \texttt{layer\_14/width\_16k/average\_l0\_84}
    \item \texttt{layer\_15/width\_16k/average\_l0\_78}
    \item \texttt{layer\_16/width\_16k/average\_l0\_78}
    \item \texttt{layer\_17/width\_16k/average\_l0\_77}
    \item \texttt{layer\_18/width\_16k/average\_l0\_74}
    \item \texttt{layer\_19/width\_16k/average\_l0\_73}
    \item \texttt{layer\_20/width\_16k/average\_l0\_71}
    \item \texttt{layer\_21/width\_16k/average\_l0\_70}
    \item \texttt{layer\_22/width\_16k/average\_l0\_72}
    \item \texttt{layer\_23/width\_16k/average\_l0\_75}
    \item \texttt{layer\_24/width\_16k/average\_l0\_73}
    \item \texttt{layer\_25/width\_16k/average\_l0\_116}
\end{itemize}

\subsubsection{Data Cleaning and Preprocessing}
All data used for fine-tuning and experiments are identical to the multilingual dataset described in Section~\ref{2.1}.
\paragraph{Data Extraction:} 
\begin{itemize}
    \item Extracted activation data from Neuronpedia JSON files.
    \item Converted structured JSON into CSV format.
    \item Standardized sentence formatting for translation and activation analysis.
\end{itemize}

\paragraph{Extracted Features:}  
\begin{itemize}
    \item \textbf{Layer Number} – Neural network layer reference.
    \item \textbf{Index Number} – Specifies neuron index.
    \item \textbf{Sentence\_en} – Extracted English sentences (activation threshold $\geq$ 0.80 of the max activation).
    \item \textbf{MaxValue\_en} – Maximum activation score per sentence.
    \item \textbf{phraseWindow\_en} – A \(7\)-token context window, consisting of the highest activation score token along with the \(3\) preceding and \(3\) following tokens.
    \item \textbf{avgMaxValue\_en} – Average activation score for English.
\end{itemize}

\subsubsection{Translation Process}

\paragraph{Method:} 
Sentences from \texttt{phraseWindow\_en} were translated into nine additional languages using Helsinki-NLP models.

\paragraph{Language Groups:}
\begin{itemize}
    \item \textbf{High-Resource Languages:}
    \begin{itemize}
        \item \texttt{phraseWindow\_en} – English
        \item \texttt{phraseWindow\_zh} – Chinese
        \item \texttt{phraseWindow\_ru} – Russian
        \item \texttt{phraseWindow\_es} – Spanish
        \item \texttt{phraseWindow\_it} – Italian
    \end{itemize}
    \item \textbf{Medium-to-Low Resource Languages:}
    \begin{itemize}
        \item \texttt{phraseWindow\_id} – Indonesian
        \item \texttt{phraseWindow\_ca} – Catalan
        \item \texttt{phraseWindow\_mr} – Marathi
        \item \texttt{phraseWindow\_ml} – Malayalam
        \item \texttt{phraseWindow\_hi} – Hindi
    \end{itemize}
\end{itemize}

\section{Fine-Tuning Setup and Evaluation}

\subsection{Fine-Tuning Configuration}

To assess the impact of activation alignment on multilingual performance, we fine-tune Gemma-\(2\)-\(2\)B using LoRA-based adaptation. The fine-tuning process targets activation disparities across multiple languages, with specific configurations as follows:

\paragraph{All Languages (Targeting Layer 20)}
\begin{itemize}
    \item Data size: \(4,000\) samples
    \item Regularization weight ($\alpha$): \(1.00\)
    \item Iterations: \(2\)
    \item Target layer: \(20\)
    \item Fine-tuned layers: \(0-20\)
\end{itemize}
The target layer refers to the layer at which we aim to reduce activation disparities between English and other languages. In this setting, we include phrases from all nine languages during fine-tuning, without designating a specific target language. The objective is to achieve broad alignment across the multilingual set. The dataset used for fine-tuning is the same as the one described in Section~\ref{2.1}.\\
Layer 20 is specifically selected to allow fine-tuning of as many layers as possible while avoiding disruption to the uppermost layers responsible for text generation.

\subsection{Benchmarking with Okapi’s Framework}

Fine-tuned models were evaluated on ARC-Challenge, HellaSwag, and MMLU using Okapi’s framework, ensuring consistency with Gemma-\(2\)-\(2\)B’s technical report evaluation settings.

\begin{itemize}
    \item \textbf{ARC-Challenge (10-shot)}: \(2,497\) samples per language
    \item \textbf{HellaSwag (10-shot)}: \(6,019\) samples per language
    \item \textbf{MMLU (5-shot)}: \(12,414\) samples per language
\end{itemize}
Each dataset was standardized to maintain comparability across languages:
\begin{itemize}
    \item \textbf{ARC-Challenge}: Common question IDs retained across languages.
    \item \textbf{HellaSwag}: Preserved identical samples per language.
    \item \textbf{MMLU}: Truncated to ensure consistency across languages.
\end{itemize}
This setup follows the Hugging Face leaderboard evaluation protocol for Gemma-\(2\)-\(2\)B.

\section{Experimental Results}

\subsection{Residual Activation Similarities Across Languages}
All pairwise cosine similarity results are available in our GitHub repository under the \textit{04 Residual Activation Calculations \& Analysis} directory. While our paper focuses on similarities between English and nine other languages, the repository contains full similarity matrices for all language pairs.

\subsection{Activation Disparities Across Layers}

To quantify the differences in activation between high-resource and medium-to-low resource languages, we measure activation differences across all \(26\) layers of Gemma-\(2\)-\(2\)B. Table~\ref{tab:activation_gap} presents these differences, showing a peak gap of \(26.27\%\) at layer \(6\). Although the gap reduces in deeper layers, a persistent difference remains, highlighting structural disparities in multilingual representation learning.

\begin{table}[t]
\centering
\begin{tabular}{c|c}
\hline
Layer & Activation Difference (\%) \\
\hline
\(0\) & \(18.06\) \\
\(1\) & \(21.31\) \\
\(2\) & \(21.82\) \\
\(3\) & \(23.59\) \\
\(4\) & \(24.89\) \\
\(5\) & \(25.77\) \\
\(6\) & \(26.27\) \\
\(7\) & \(25.55\) \\
\(8\) & \(24.84\) \\
\(9\) & \(23.41\) \\
\(10\) & \(23.59\) \\
\(11\) & \(22.25\) \\
\(12\) & \(21.60\) \\
\(13\) & \(20.52\) \\
\(14\) & \(20.49\) \\
\(15\) & \(22.25\) \\
\(16\) & \(21.54\) \\
\(17\) & \(21.68\) \\
\(18\) & \(22.64\) \\
\(19\) & \(23.67\) \\
\(20\) & \(23.15\) \\
\(21\) & \(22.64\) \\
\(22\) & \(23.20\) \\
\(23\) & \(22.80\) \\
\(24\) & \(20.75\) \\
\(25\) & \(19.89\) \\
\hline
\end{tabular}
\caption{Activation differences between high-resource and medium-to-low resource languages across all layers.}
\label{tab:activation_gap}
\end{table}

\subsection{Benchmark Performance Across Languages}

Tables~\ref{tab:base_benchmark_arc}, \ref{tab:base_benchmark_hellaswag}, and \ref{tab:base_benchmark_mmlu} present the accuracy results for high-resource and medium-to-low resource languages across different tasks. The results show that high-resource languages consistently outperform medium-to-low resource languages across all benchmarks, reinforcing that weaker activations in feature space lead to weaker performance in downstream tasks.
\begin{table}[t]
\centering
\begin{tabular}{lc}
    \hline
    Language & ARC-Challenge (\%) \\
    \hline
    \multicolumn{2}{c}{High-Resource Languages} \\
    \hline
    English & \(53.67\) \\
    Chinese & \(42.22\) \\
    Russian & \(37.90\) \\
    Spanish & \(45.07\) \\
    Italian & \(43.91\) \\
    \hline
    \multicolumn{2}{c}{Medium-to-Low Resource Languages} \\
    \hline
    Indonesian & \(40.14\) \\
    Catalan & \(38.71\) \\
    Marathi & \(26.16\) \\
    Malayalam & \(26.34\) \\
    Hindi & \(30.47\) \\
    \hline
\end{tabular}
\caption{ARC-Challenge benchmark accuracy for Gemma-\(2\)-\(2\)B.}
\label{tab:base_benchmark_arc}
\end{table}
\begin{table}[t]
\centering
\begin{tabular}{lc}
    \hline
    Language & HellaSwag (\%) \\
    \hline
    \multicolumn{2}{c}{High-Resource Languages} \\
    \hline
    English & \(76.23\) \\
    Chinese & \(54.98\) \\
    Russian & \(54.54\) \\
    Spanish & \(62.95\) \\
    Italian & \(57.72\) \\
    \hline
    \multicolumn{2}{c}{Medium-to-Low Resource Languages} \\
    \hline
    Indonesian & \(55.92\) \\
    Catalan & \(50.24\) \\
    Marathi & \(30.99\) \\
    Malayalam & \(29.44\) \\
    Hindi & \(39.77\) \\
    \hline
\end{tabular}
\caption{HellaSwag benchmark accuracy for Gemma-\(2\)-\(2\)B.}
\label{tab:base_benchmark_hellaswag}
\end{table}
\begin{table}[h]
\centering
\begin{tabular}{lc}
    \hline
    Language & MMLU (\%) \\
    \hline
    \multicolumn{2}{c}{High-Resource Languages} \\
    \hline
    English & \(49.15\) \\
    Chinese & \(40.71\) \\
    Russian & \(40.22\) \\
    Spanish & \(44.57\) \\
    Italian & \(43.70\) \\
    \hline
    \multicolumn{2}{c}{Medium-to-Low Resource Languages} \\
    \hline
    Indonesian & \(42.30\) \\
    Catalan & \(42.16\) \\
    Marathi & \(31.54\) \\
    Malayalam & \(29.44\) \\
    Hindi & \(35.85\) \\
    \hline
\end{tabular}
\caption{MMLU benchmark accuracy for Gemma-\(2\)-\(2\)B.}
\label{tab:base_benchmark_mmlu}
\end{table}

\subsection{Activation Improvements at Layer 20 Across Languages}

Fine-tuning at layer \(20\) resulted in activation increases across all ten languages. Table~\ref{tab:activation_improvement} presents the activation improvements for each language post-fine-tuning, while Table~\ref{tab:english_activation_retention} reports the retention of English activations after fine-tuning. 
\begin{table}[t]
\centering
\begin{tabular}{lc}
    \hline
    Language & Activation Improvement (\%) \\
    \hline
    Malayalam & \(87.69\) \\
    Hindi & \(86.32\) \\
    Marathi & \(93.87\) \\
    Catalan & \(63.22\) \\
    Indonesian & \(56.58\) \\
    Italian & \(43.48\) \\
    Spanish & \(43.71\) \\
    Russian & \(70.08\) \\
    Chinese & \(66.93\) \\
    \hline
\end{tabular}
\caption{Activation improvements across ten languages after fine-tuning.}
\label{tab:activation_improvement}
\end{table}
\begin{table}[h]
\centering
\begin{tabular}{lc}
    \hline
    Language & English Activation Retention (\%) \\
    \hline
    Malayalam & \(90.40\) \\
    Hindi & \(91.90\) \\
    Marathi & \(92.81\) \\
    Catalan & \(92.96\) \\
    Indonesian & \(90.46\) \\
    Italian & \(91.82\) \\
    Spanish & \(91.02\) \\
    Russian & \(90.70\) \\
    Chinese & \(93.17\) \\
    \hline
\end{tabular}
\caption{English activation retention after fine-tuning.}
\label{tab:english_activation_retention}
\end{table}

\subsection{Activation Improvements for Malayalam Across Various Layers}
\begin{figure}[ht]
    \centering
    \includegraphics[width=1\linewidth]{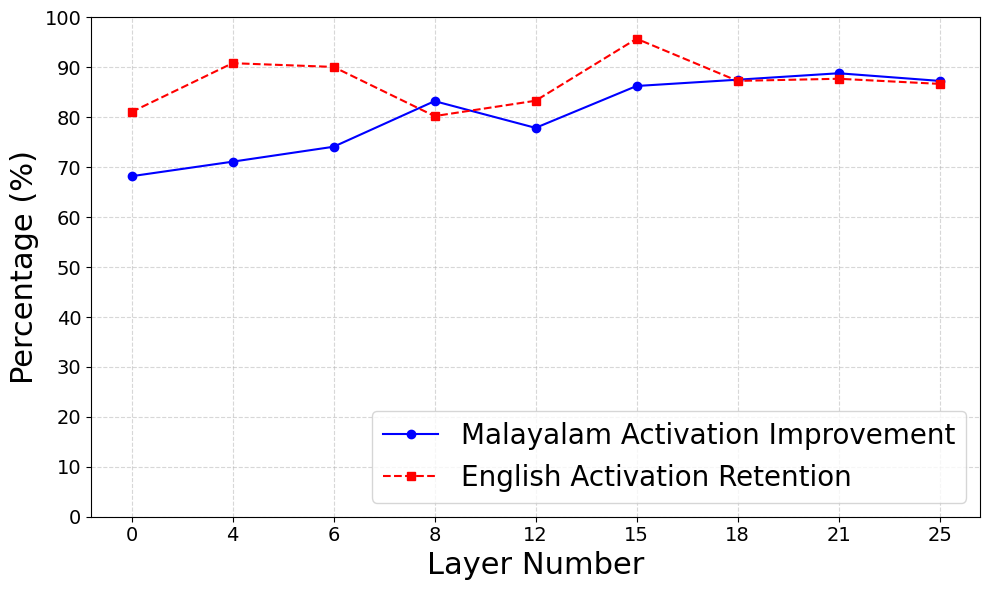}
\caption{Activation alignment for Malayalam across various layers.}
\label{fig:malayalam_all_layers}
\end{figure}

Malayalam, which initially exhibited the largest activation disparity, demonstrated substantial improvements in activation levels across all layers after fine-tuning. As shown in Figure~\ref{fig:malayalam_all_layers}, the activation differences are reduced across the layers that were randomly selected for the experiment, aligning Malayalam more closely with high-resource languages.\\
\noindent
To quantify these improvements, Table~\ref{tab:activation_improvement_ml} presents the percentage increase in activation values for Malayalam across different layers. For efficiency, we randomly selected a subset of layers to serve as representative checkpoints. This selection ensures that improvements are observed across early, middle, and deep layers without introducing bias in specific regions of the model.\\
\begin{table}[t]
\centering
\begin{tabular}{lc}
    \hline
    Layer & Malayalam Activation Improvement (\%) \\
    \hline
    \(0\)  & \(68.21\) \\
    \(4\)  & \(71.12\) \\
    \(6\)  & \(74.11\) \\
    \(8\) & \(83.25\) \\
    \(12\) & \(77.89\) \\
    \(15\) & \(86.28\) \\
    \(18\) & \(87.54\) \\
    \(21\) & \(88.83\) \\
    \(25\) & \(87.30\) \\
    \hline
\end{tabular}
\caption{Layer-wise activation improvement for Malayalam after fine-tuning.}
\label{tab:activation_improvement_ml}
\end{table}
\noindent
While fine-tuning successfully increased Malayalam's activation levels, it is also important to assess whether English activations were preserved. Table~\ref{tab:activation_retention_en} reports the retention rates of English activations across the same layers. Notably, English activations remained largely stable, with retention exceeding \(80\%\) in all layers and peaking at \(95.75\%\) in layer \(15\). This indicates that improvements in Malayalam did not come at the cost of significant degradation in English performance.\\
\begin{table}[h]
\centering
\begin{tabular}{lc}
    \hline
    Layer & English Activation Retention (\%) \\
    \hline
    \(0\)  & \(81.08\) \\
    \(4\)  & \(90.85\) \\
    \(6\)  & \(90.10\) \\
    \(8\)  & \(80.25\) \\
    \(12\) & \(83.33\) \\
    \(15\)& \(95.75\) \\
    \(18\) & \(87.30\) \\
    \(21\) & \(87.73\) \\
    \(25\) & \(86.70\) \\
    \hline
\end{tabular}
\caption{Layer-wise English activation retention after fine-tuning. The selected layers serve as checkpoints to assess consistency across different depths.}
\label{tab:activation_retention_en}
\end{table}
\noindent
These results demonstrate that our fine-tuning approach, guided by the loss function incorporating activation disparity, successfully reduces the activation gap for Malayalam across layers while preserving English activations. The combination of Figure~\ref{fig:malayalam_all_layers} and Tables~\ref{tab:activation_improvement_ml} and~\ref{tab:activation_retention_en} provides a comprehensive validation of our loss formulation's impact across multiple layers in Gemma-\(2\)-\(2\)B model.

\subsection{Model Performance After Fine-tuning for Malayalam Across All Benchmarks}

\begin{figure}[h]
\centering
\includegraphics[width=1\linewidth]{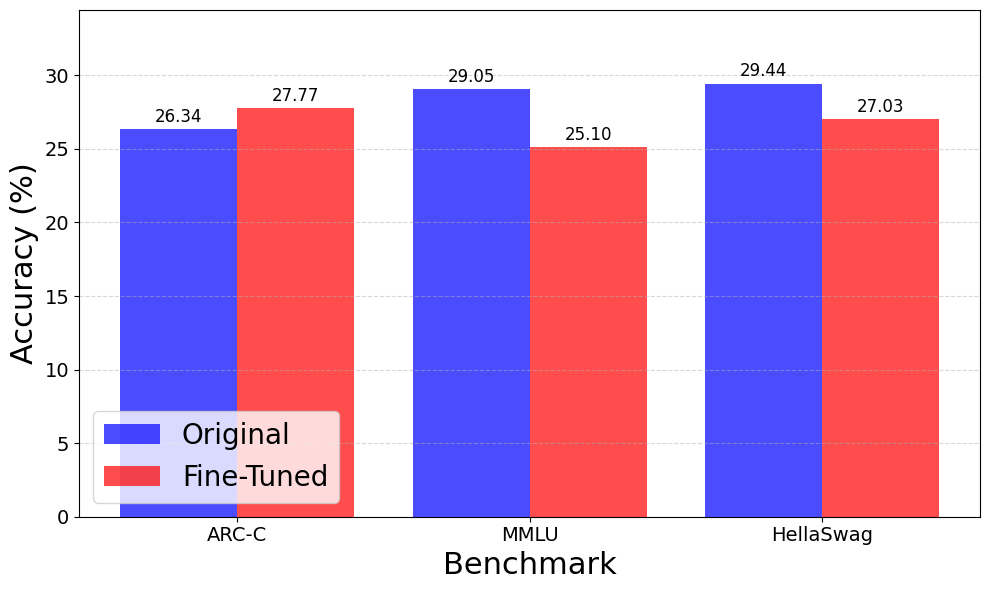}
\caption{Benchmark performance across all tasks for Malayalam after fine-tuning.}
\label{fig:benchmark_results_malayalam}
\end{figure}

While fine-tuning significantly improved activation levels, its impact on benchmark performance was inconsistent. As shown in Figure~\ref{fig:benchmark_results_malayalam}, the model exhibited varying degrees of performance changes across ARC-Challenge (ARC-C), MMLU, and HellaSwag.
Table~\ref{tab:malayalam_layer20} quantifies these results. Notably, ARC-C showed a performance gain, increasing \(1.44\%\) from \(26.34\%\) to \(27.77\%\) after fine-tuning. This suggests that the model may have improved in reasoning-based tasks, potentially benefiting from better token representations at deeper layers.
\noindent
However, performance in MMLU and HellaSwag declined post fine-tuning. MMLU, which requires broad factual knowledge, dropped \(3.95\%\) from \(29.05\%\) to \(25.10\%\), while HellaSwag, which involves commonsense reasoning, saw a \(2.41\%\) decrease from \(29.44\%\) to \(27.03\%\). This inconsistency indicates that while fine-tuning aligned activations for Malayalam, it did not translate uniformly into downstream improvements across all tasks. The decline in MMLU suggests a potential trade-off where fine-tuning optimizes some linguistic representations at the cost of factual recall.
\begin{table}[t]
\centering
\begin{tabular}{lcl}
    \hline
    Benchmark & Score Type & Accuracy (\%) \\
    \hline
    ARC-C & Original & \(26.34\) \\
          & Fine-Tuned & \(27.77\) \\
    \hline
    MMLU & Original & \(29.05\) \\
         & Fine-Tuned & \(25.10\) \\
    \hline
    HellaSwag & Original & \(29.44\) \\
              & Fine-Tuned & \(27.03\) \\
    \hline
\end{tabular}
\caption{Performance of Malayalam at layer \(20\) across different benchmarks before and after fine-tuning.}
\label{tab:malayalam_layer20}
\end{table}

\noindent
These results highlight that activation alignment does not equally translate into higher benchmark scores. The observed improvement in ARC-C suggests that fine-tuning enhanced reasoning abilities, but the decline in MMLU and HellaSwag implies that certain knowledge-intensive tasks may have been negatively impacted. Future work could investigate whether task-specific fine-tuning strategies or balancing activation alignment across layers can lead to more consistent performance gains across all benchmarks.

\section{Cross-Linguistic Activation Disparities}
\begin{figure*}[ht]
  \centering
  \includegraphics[width=1.0\linewidth]{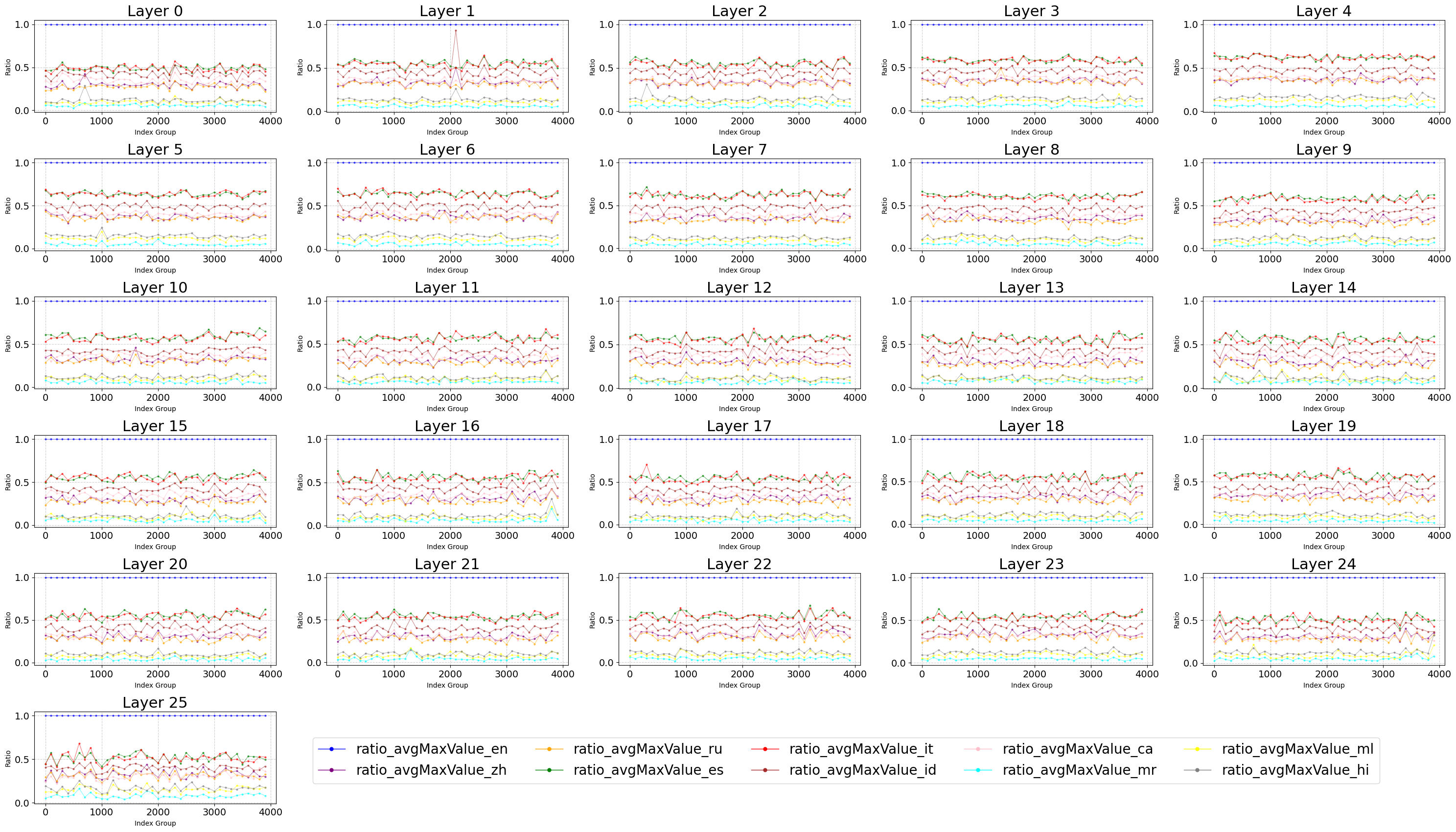} 
  \caption{Activation ratios across all layers and 4000 feature indices, comparing activation values for each language relative to English.}
  \label{fig:overall_differences_all_languages}
\end{figure*} 

\begin{table}[ht]
\centering
\begin{tabular}{lcc}
    \hline
    Language & Mean & Std \\
    \hline
    Chinese     & \(0.33\) & \(0.32\) \\
    Russian     & \(0.30\) & \(0.31\) \\
    Spanish     & \(0.57\) & \(0.34\) \\
    Italian     & \(0.56\) & \(0.36\) \\
    Indonesian  & \(0.43\) & \(0.36\) \\
    Catalan     & \(0.35\) & \(0.33\) \\
    Marathi     & \(0.05\) & \(0.17\) \\
    Malayalam   & \(0.10\) & \(0.24\) \\
    Hindi       & \(0.12\) & \(0.25\) \\
    \hline
\end{tabular}
\caption{Mean and standard deviation of activation ratios with English across all 26 layers for each language.}
\label{tab:mean_std}
\end{table}

\noindent
From Figure~\ref{fig:overall_differences_all_languages}, the results show a systematic disparity, with all languages consistently receiving lower
activations than English. Table~\ref{tab:mean_std} confirms that this disparity persists across all layers. Additionally, we observe variability across languages:

\noindent
1. Malayalam and Marathi exhibit the largest activation gaps, confirming that these languages are underrepresented in feature space.

\noindent
2. High-resource languages such as Chinese, Russian, and Spanish still show activation gaps, though smaller than medium-to-low resource languages.

\noindent
These findings reinforce that activation disparities are a structural phenomenon in multilingual LLMs, where medium-to-low resource languages systematically receive fewer learned representations. This under-representation likely contributes to performance degradation on multilingual benchmarks.

\begin{figure*}[ht]
  \centering
  \includegraphics[width=1.0\linewidth]{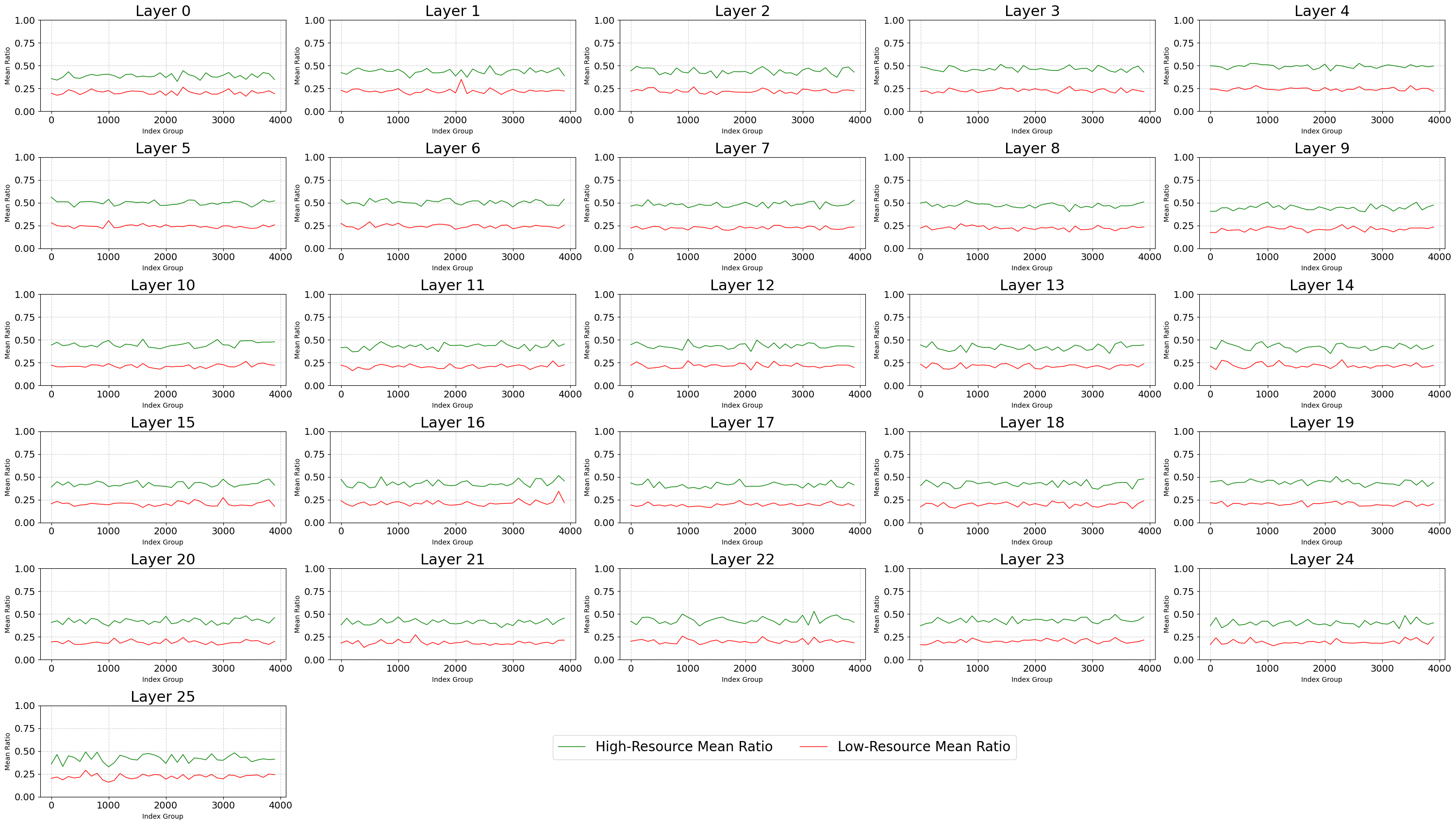} 
  \caption{Comparison of activation differences between high-resource and medium-to-low resource language groups.}
  \label{fig:overall_difference_high_vs_low}
\end{figure*} 

\noindent
From Figure~\ref{fig:overall_difference_high_vs_low}, the results highlight a persistent activation gap, where high-resource languages receive consistently higher activations than medium-to-low resource languages across all layers. This trend suggests systematic differences in how multilingual models process linguistic features, which can be examined through the following key observations:

\noindent
1. Early layers show the largest gap, reinforcing that medium-to-low resource languages struggle to activate neurons encoding foundational linguistic features.

\noindent
2. Disparities persist in deeper layers, indicating that even though later representations are more abstract and shared across languages, medium-to-low resource languages still do not fully align with high-resource languages in feature space.

\noindent
3. The separation between high-resource and medium-to-low resource languages remains stable, suggesting that multilingual models inherently prioritize high-resource languages at all levels of representation.

\noindent
These findings further support the hypothesis that activation disparities contribute to cross-linguistic performance gaps, as medium-to-low resource languages remain underrepresented in the model’s learned feature space.

\begin{figure}[h]
\centering
\includegraphics[width=1\linewidth]{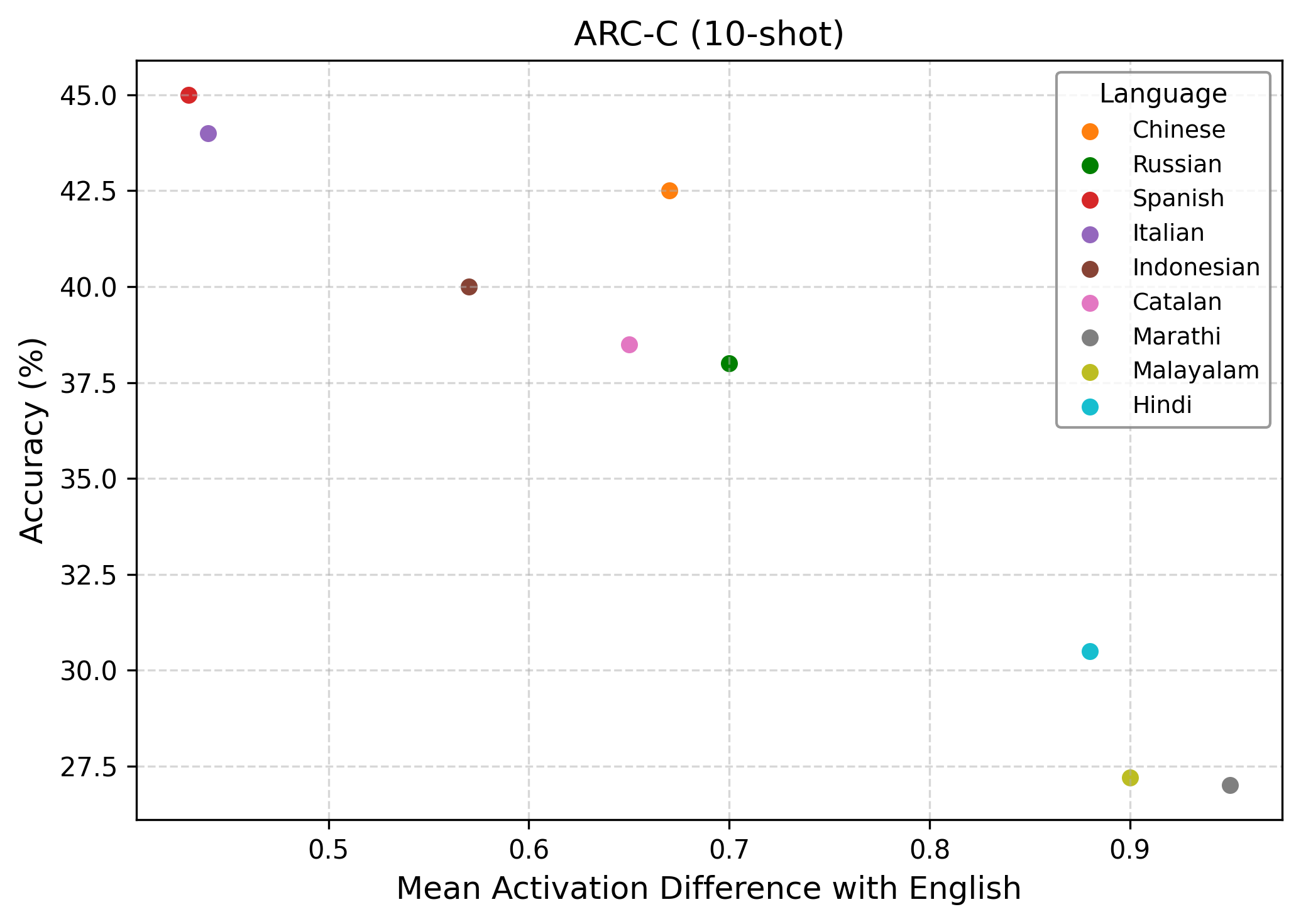}
\caption{Correlation between activation difference with English and ARC-C accuracy across languages. Pearson’s correlation coefficient: \( r = -0.95 \).}
\label{fig:figure_7}
\end{figure}

\begin{figure}[h]
\centering
\includegraphics[width=1\linewidth]{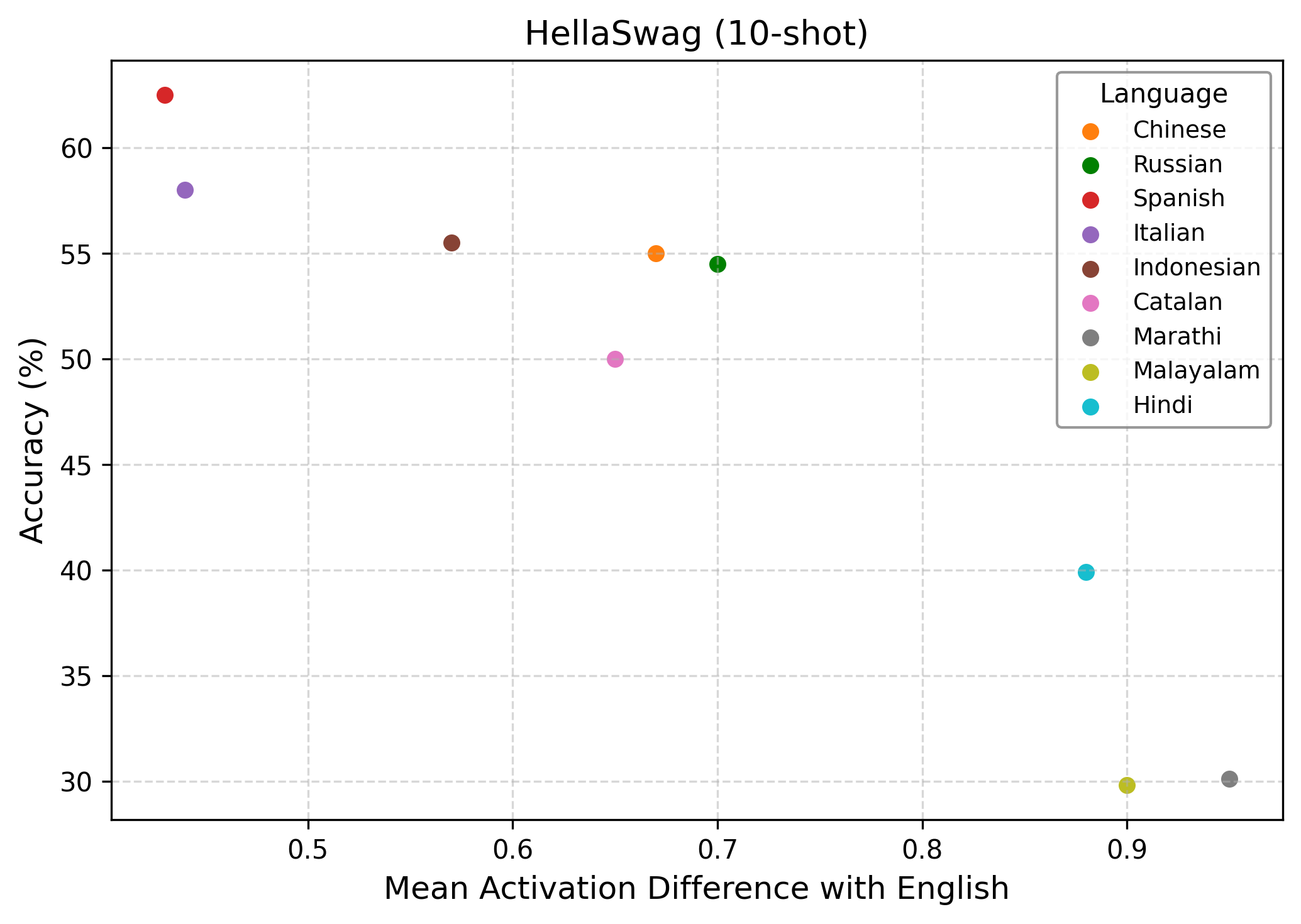}
\caption{Correlation between activation difference with English and HellaSwag accuracy across languages. Pearson’s correlation coefficient: \( r = -0.93 \).}
\label{fig:figure_8}
\end{figure}

\begin{figure}[h]
\centering
\includegraphics[width=1\linewidth]{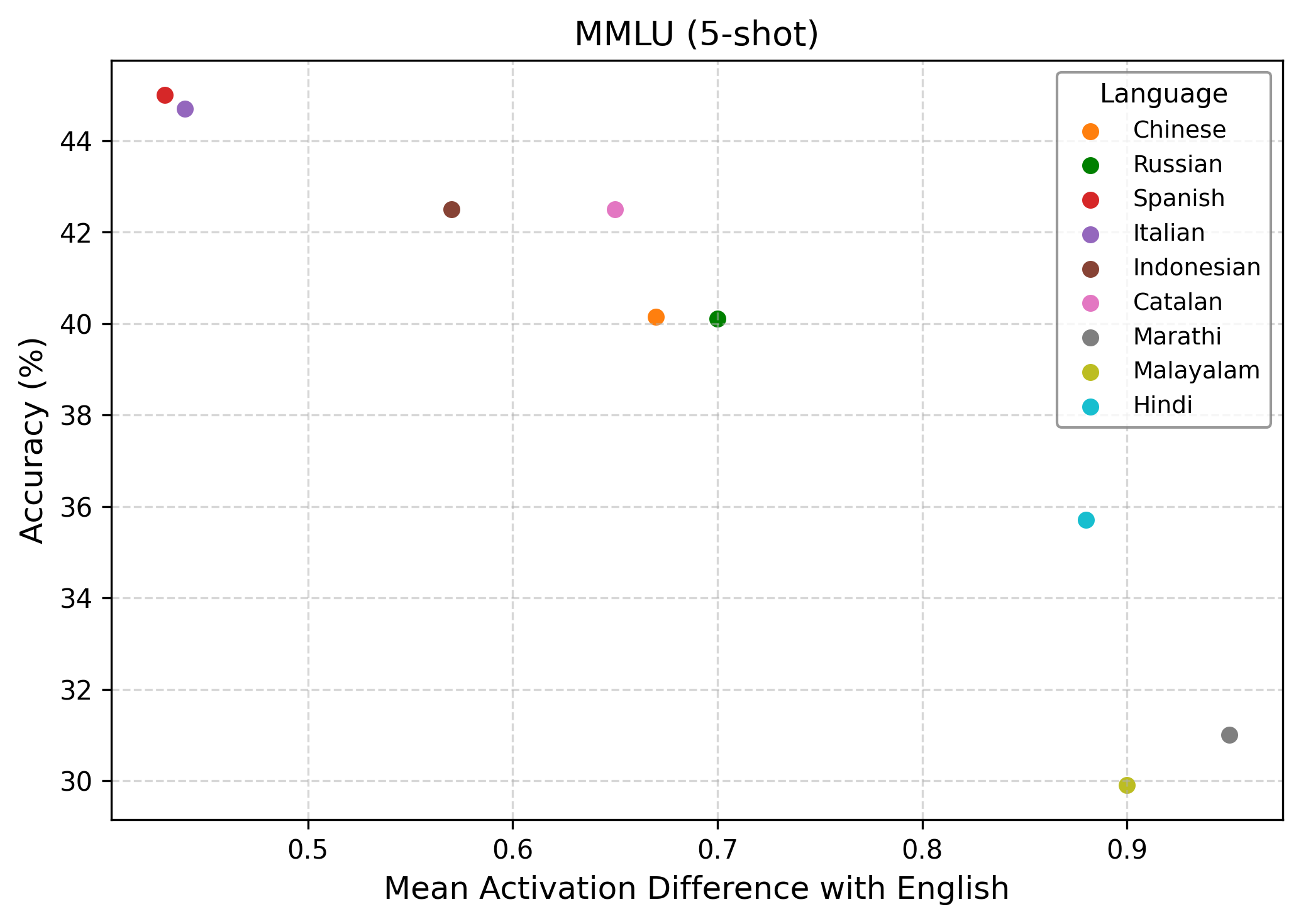}
\caption{Correlation between activation difference with English and MMLU accuracy across languages. Pearson’s correlation coefficient: \( r = -0.95 \).}
\label{fig:figure_9}
\end{figure}
\end{document}